\documentclass[conference, 10pt]{IEEEtran}
\IEEEoverridecommandlockouts
\usepackage{cite}
\usepackage{amsmath,amssymb,amsfonts}
\usepackage{algorithmic}
\usepackage{caption}
\usepackage{subcaption}
\usepackage{graphicx}
\usepackage{textcomp}
\usepackage{xcolor}
\usepackage{dsfont}
\usepackage{balance}
\usepackage{subcaption}
\usepackage{lipsum}
\def\BibTeX{{\rm B\kern-.05em{\sc i\kern-.025em b}\kern-.08em
    T\kern-.1667em\lower.7ex\hbox{E}\kern-.125emX}}

\begin{document}

\title{
Goal-oriented Communications based on Recursive Early Exit Neural Networks
\thanks{This work has been supported by the SNS JU project 6G-GOALS under the EU’s Horizon program Grant Agreement No 101139232, by Sapienza grant RG123188B3EF6A80 (CENTS), by European Union under the Italian National Recovery and Resilience Plan of NextGenerationEU, partnership on Telecommunications of the Future (PE00000001 - program RESTART), and by the ANR under the France 2030 program, grant "NF-NAI: ANR-22-PEFT-0003".}
}

\author{Jary Pomponi\IEEEauthorrefmark{2}\IEEEauthorrefmark{3}\textsuperscript{\textsection}, 
Mattia Merluzzi\IEEEauthorrefmark{6}\textsuperscript{\textsection}, 

\IEEEauthorblockN{Alessio Devoto\IEEEauthorrefmark{1}\IEEEauthorrefmark{3}, Mateus Pontes Mota\IEEEauthorrefmark{6},\\  Paolo Di Lorenzo\IEEEauthorrefmark{2}\IEEEauthorrefmark{3} and Simone Scardapane\IEEEauthorrefmark{2}\IEEEauthorrefmark{3}}
\IEEEauthorblockA{
\IEEEauthorrefmark{1}DIAG Department, Sapienza University of Rome, Via Ariosto 25, 00185, Rome, Italy}
\IEEEauthorblockA{
\IEEEauthorrefmark{3}Consorzio Nazionale Interuniversitario per le Telecomunicazioni, Viale G.P. Usberti, 181/A, 43124, Parma, Italy}
\IEEEauthorblockA{
\IEEEauthorrefmark{2}DIET Department, Sapienza University of Rome, Via Eudossiana 18, 00184, Rome, Italy\\ Emails: \emph{\{firstname.lastname\}@uniroma1.it}}
\IEEEauthorblockA{
\IEEEauthorrefmark{6}CEA-Leti, Université Grenoble Alpes, F-38000 Grenoble, France \\ Emails: \emph{\{firstname.lastname\}@cea.fr}}
\vspace{-.7cm}}

\maketitle

\begingroup\renewcommand\thefootnote{\textsection}
\footnotetext{Equal contribution}
\endgroup

\begin{abstract}
This paper presents a novel framework for goal-oriented semantic communications leveraging recursive early exit models. The proposed approach is built on two key components. First, we introduce an innovative early exit strategy that dynamically partitions computations, enabling samples to be offloaded to a server based on layer-wise recursive prediction dynamics that detect samples for which the confidence is not increasing fast enough over layers. Second, we develop a Reinforcement Learning-based online optimization framework that jointly determines early exit points, computation splitting, and offloading strategies, while accounting for wireless conditions, inference accuracy, and resource costs. Numerical evaluations in an edge inference scenario demonstrate the method's adaptability and effectiveness in striking an excellent trade-off between performance, latency, and resource efficiency.
\end{abstract}

\vspace{.15cm}
\begin{IEEEkeywords}
Goal-oriented semantic communications, early exit neural networks, adaptive computation, resource allocation. \vspace{-.3cm}
\end{IEEEkeywords}

\section{Introduction}

Recent years have witnessed a surge of interest in goal-oriented semantic communications \cite{strinati2024goal}, which transcend traditional bit-centric metrics commonly employed in system design and optimization, emphasizing the recovery of the meaning encoded within the transmitted bits and/or ensuring the effective completion of the tasks that drive the information exchange. A fundamental enabler of this research direction is artificial intelligence (AI) and, specifically, deep learning (DL) models that can extract from data only the features that are strictly relevant and must be communicated \cite{strinati2024goal,di2023goal}. However, current state-of-the-art DL models demand substantial memory, energy, and computation \cite{qu2020deep}, making them unsuitable for communication systems without tailored design. Their requirements often clash with the device-server paradigm, where devices have limited computational and energy resources, preventing full execution on-device. Offloading computations to the cloud or edge servers, while feasible, introduces challenges such as increased end-to-end latency, potential privacy risks from transmitting raw data, and susceptibility to noise and errors in communication channels \cite{8763885,9116699}.

A hybrid approach can address the aforementioned limitations by determining the optimal strategy for offloading data from the device to the server. This decision is strongly influenced by the dynamic availability of computational resources on the device and at (edge) servers, as well as by the varying conditions of the wireless channel. Such an interplay between communication and computation is a hallmark of next-generation wireless networks, which serve as key enablers and potentially efficient platforms for delivering and accessing Artificial Intelligence (AI) as a service \cite{MerluzziHexa23}. The final objective is to find the partition that best balances end-to-end (loop \cite{Ana21}) latency, communication and computation costs, while guaranteeing a minimum target level of performance. 
However, fixing the partitioning point a priori is sub-optimal due to the changing nature of the whole system. A better solution would be to learn and adapt the splitting point over time \cite{9700550, Binucci24}.



In this context, early exiting is a powerful technique to avoid computing across the whole DL model for all input data, by retrieving the result after an intermediate layer of the neural network \cite{teerapittayanon2016branchynet}. Several early exit (EE) points can be placed across a DL model in order to gain degrees of freedom in optimizing splitting point and EE in a dynamic goal-oriented fashion, based on time-varying resource availability and wireless channel quality. Such architectures must be designed by taking into account the communication burden of transmitting intermediate results for further processing at a selected edge server, and to reduce the payload size as much as possible, without over-sacrificing performance. From a communication perspective, the optimization of the splitting/early exit choice, depending on inference performance target and communication costs (e.g., energy consumption and bandwidth usage), has been shown to be the best solution \cite{Bullo23, Jankowski24,  Colocrese24}.

The aim of this paper is twofold: (i) we propose a novel early exit approach that not only halts samples that can be solved locally on the device, but also implements a dynamic partition approach in which samples can be sent to the server model at any step of the device-side computation, using a criterion based on how fast the prediction changes from one layer to another; (ii) we introduce an online optimization that dynamically decides early exit, splitting point, and offloading strategy, considering wireless channel conditions, inference performance, and communication/computation costs. Differently from the existing literature \cite{Bullo23, Jankowski24,  Colocrese24}, the online optimization is based on a Reinforcement Learning procedure that takes into consideration  communication, computation, and learning aspects in a joint manner. Also, the proposed early exit model is specifically designed for the considered goal-oriented communication setup. Numerical results illustrate the performance of the proposed approach in an edge inference scenario, and its flexibility in balancing inference performance, delay, communication and computation costs.

\section{Recursive Early Exit Neural Networks}\label{sec:EE_methodology}
%

A generic neural network for classification can be represented as a composition of $b$ sequential blocks followed by a classification layer $c_b$:
\begin{equation}\label{eq:NN_model}
    f_b(x) = c_b \circ \mathrm{l}_{b} \circ \mathrm{l}_{b-1} \circ \cdots \circ \mathrm{l}_{1}(x),     
\end{equation}
where $x$ is the input data, $\mathrm{l}_i$ represents the mapping of the $i$-th layer, and $c_b$ is the final classification layer. In the context of EE models, the goal is to enable the model to classify samples at intermediate layers $i < b$. To achieve this, auxiliary classifiers are attached and trained at specific inner layers, in addition to the final classifier $c_b$. Let the set of indices of hidden layers with attached exits be $\mathcal{I}$, where $i \in \mathcal{I}$ and $i < b$. The $i$-th exit is defined as:
\begin{equation}\label{eq:gen_EE_model}
    f_i(x) = c_i \circ \mathrm{l}_i \circ \mathrm{l}_{i-1} \circ \cdots \circ \mathrm{l}_{1}(x),
\end{equation}
where $c_i$ is the classifier attached to the $i$-th layer. Each $c_i$ consists of a combination of a convolutional module and a linear layer and can be decomposed as $c_i(x) = p_i \circ e_i(x)$. Here, $e_i$ processes the inner activations of the model, and $p_i$ maps the processed features to the output prediction vector. The simplest approach to train the neural network model in \eqref{eq:NN_model}-\eqref{eq:gen_EE_model} is to jointly optimize all classifiers by minimizing a weighted sum of their individual loss functions, see, e.g., \cite{teerapittayanon2016branchynet}.



\noindent \textbf{Recursive Architecture.} Similarly to \cite{probreint,wolczyk2021zero}, we structure our model as a recursive one, in which the predictions at a given exit are created by recursively combining all the predictions up to the current point. Specifically, our model gives a probability for each class independently, so that $f_i^j(x) \in [0, 1]$, with $j \in \mathcal{C}$, and $\mathcal{C}$ being the set of classes. To build such predictions recursively, at a generic early exit layer $i$, we combine the intermediate output of it, $e_i(x)$, with the previous prediction $f_{i-1}(x)$ to create the vector of the current predictions $ f_i(x) = c_i(f_{i-1}(x), e_{i}(x))$, where $c_i(\cdot)$ is a classification function that combines the input values to create the current prediction. 
In particular, given a hidden layer $i > 1$, the vector of its inner activations $e_{i}(x)$ is the input of two separate functions. The first function, $m^+_i(x)$, is referred to as the positive moving mass function, while the second, $m^-_i(x)$, is the negative moving mass function. Both functions consist of a linear layer followed by a sigmoid activation to constrain the output to $[0, 1]$, and are used to determine how much the prediction for each class changes between consecutive layers. Mathematically, we have:
\begin{equation*}
\label{eq:probs}
    c_i(x) = 
\begin{cases} 
     f_{1}(x) & 
     i = 1 \\
     f_{i-1}(x) + M_i(x) & 
     i \ne b 
     \\ f_{i-1}(x) + (1 - f_{i-1}(x)) \cdot f_b(x)  & i = b 
 \end{cases}
\end{equation*}
where $f_b(x)$ is assumed to be a distribution, and $M_i(x)$ is the mass that can detracted from or added to the current predictions, defined as:
\begin{align*}
    \label{eq:mass}
    M_i(x) = (1 - f_{i-1}(x))  \cdot m^+_i(e_{i}(x)) - f_{i-1}(x)  \cdot m^-_i(e_{i}(x)).
\end{align*}
The key idea is to adjust the cumulative probability by adding a fraction ($m^+_i(x)$) of the remaining probability $(1 - f_{i-1})$ and subtracting a fraction ($m^-_i(x)$) of the cumulative probability $(f_{i-1})$, requiring only the duplication of the linear layer after the exit for each moving mass function, thus minimally increasing parameters compared to the formulation in \cite{teerapittayanon2016branchynet}. 
\noindent \textbf{Training.} Given a generic training tuple $(x, y)$ and the associated predictions $f_{i}(x)$ for each classifier, we want to regularize the probabilities so that the correct one, $f^y_{i}(x)$, is larger than the other ones for each exit. 
To this aim, we first define a function returning the highest probability which is not $f^y_i$, i.e.,
 $   \text{H}_{i}(x) = \max \{f^{j}_{i}(x) \,\vert\, j \in \mathcal{C} \land j \ne y \}$, and then define our classification loss for exit $b > i \ge 1$ as: 
\begin{equation*}
\label{eq:margin_loss}
\mathcal{D}_i(x, y) = \max\left(0, \, \text{H}_{i}(x) - f^y_{i}(x) + m \right),
\end{equation*}
where $m$ is a safety confidence margin that we set to $\frac{2}{\lvert \mathcal{C} \rvert}$. Then, letting $\text{CE}$ be the cross-entropy function, the overall loss can be written as:
\begin{equation}
\label{eq:margin_loss}
\mathcal{L}_\text{mass}(x, y) = \text{CE}(f_b(x), y) + \frac{1}{b-1} \sum_{i=1}^{b-1} \mathcal{D}_i(x, y)
\end{equation}
The loss \eqref{eq:margin_loss} promotes a safe distance, given by the margin, between the probability of the correct class and the others.
%
%
%
%
%
\noindent \textbf{Halting.} Once selected a margin $m$, the computation is halted at a generic exit $i<b$ if $f^{t_1}_{i}(x) - f^{t_2}_{i}(x) > m$,
%
%
where $f^{t_1}$ and $f^{t_2}$ are the first and second-highest probability produced in output by the model. A larger margin enforces a more selective halting process, prompting the model to halt only on samples with low classification uncertainty, thereby enhancing early exit performance while reducing the number of halted samples. Ultimately, the margin serves as a tunable parameter to balance computational efficiency and accuracy.

\begin{figure*}[t!]
    \centering
    \begin{minipage}[a]{0.6\linewidth}
        \includegraphics[width=\columnwidth]{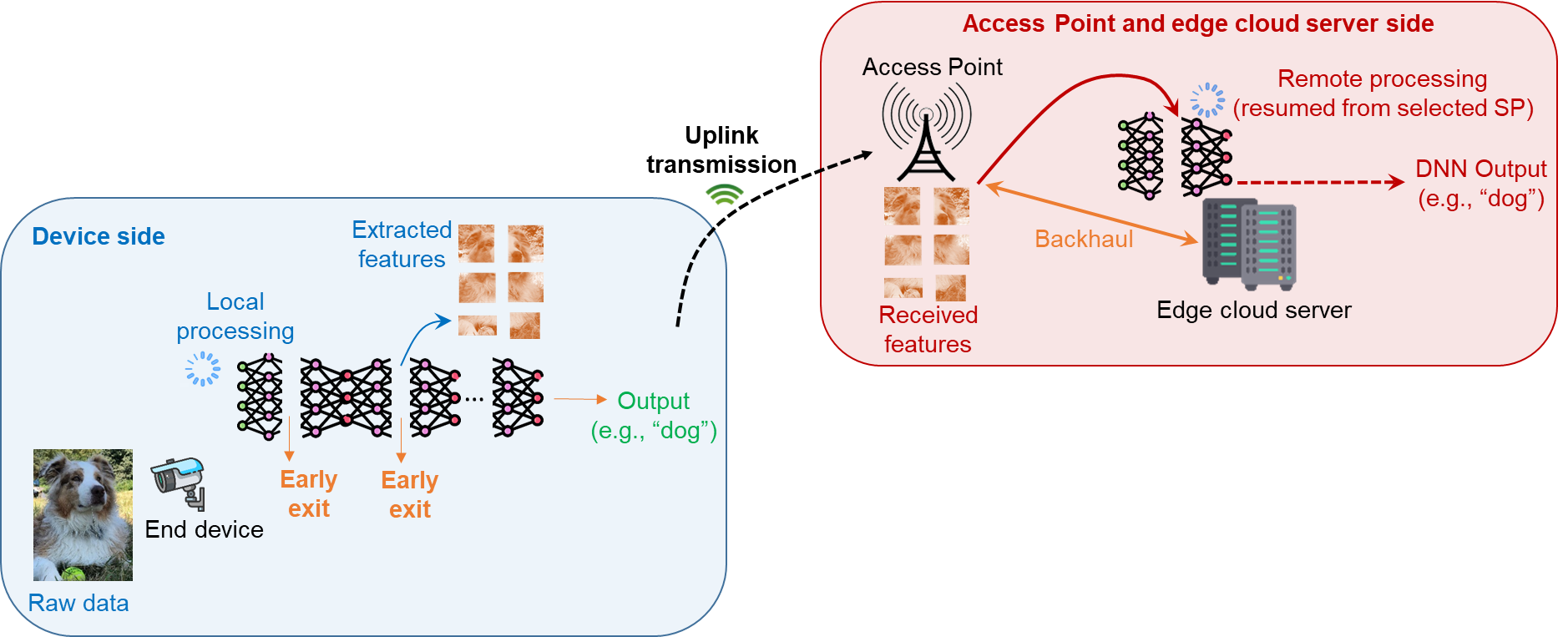}
        \caption{System scenario for EE-based goal-oriented communication.}
        \label{fig:wireless_scenario}
    \end{minipage}
    \hspace{10pt}
    \begin{minipage}[a]{0.35\linewidth}
        \includegraphics[width=\columnwidth]{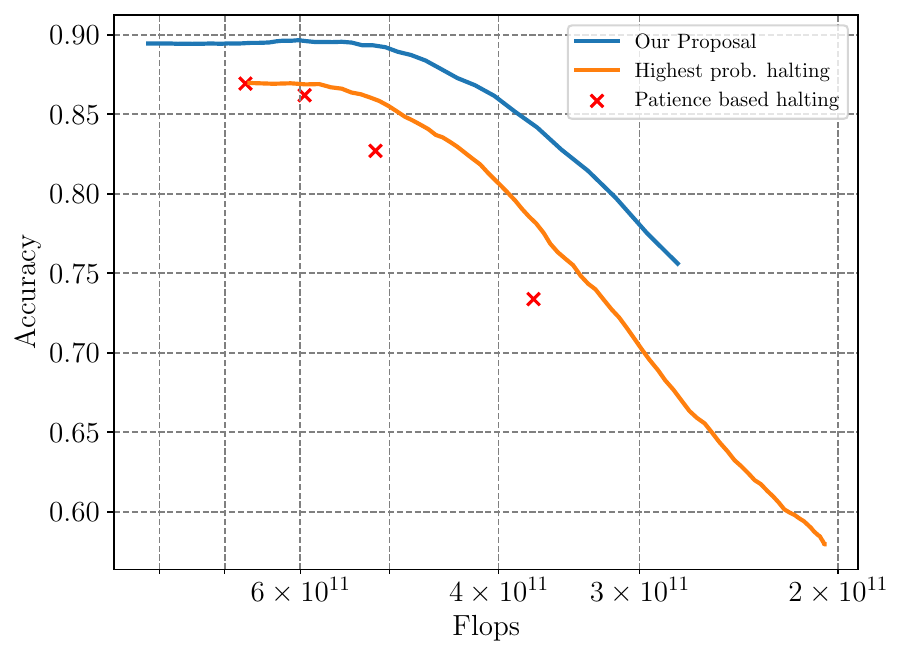}
        \caption{Accuracy-efficiency trade-off, comparing different methods.}
        \label{fig:flops}
    \end{minipage}
\end{figure*}

\noindent \textbf{Numerical results.} We train and validate a ResNet20 \cite{he2016deep} on CIFAR10 dataset. 
For each layer of the model, we place an EE after each inner block, creating 9 early exits, which are trained using the proposed approach. Figure \ref{fig:flops} shows the results of our model against two strategies, namely, "Highest probability halting" and "patience-based". Both competitors are trained as in \cite{teerapittayanon2016branchynet}, with the former halting computation when a threshold over the highest probability is reached, whereas the latter halts when a sample is classified the same for a given number of exits \cite{zhou2020bert}. 
For each approach, we test multiple thresholds, margins, or patience values. The trade-off between operations (FLOPs) and accuracy is shown in Figure \ref{fig:flops}, where we can see how our approach consistently outperforms the competitors.
\section{Early exiting and splitting over wireless}\label{EE_over_wireless}
%
%
In this section, we consider a scenario in which the described NN architecture with early exits is trained and embedded in an end device, as well as in an edge server that can take over computation whenever needed. The scenario is illustrated in Fig. \ref{fig:wireless_scenario}. Then, the architecture is available at inference time on data that are continuously collected by the end device, with enhanced degrees of freedom on where to place workload and, when possible, where to stop local computation to promote offloading, based on time-varying wireless and compute resource conditions. The device is then faced with a dynamic choice involving early exiting, computing more, or transmitting intermediate results for further processing at the edge server. To this end, relevant parameter include the size of intermediate results and the computational load. At the same time, relevant key performance indicators (KPIs) include the delay and the accuracy. We denote by $\mathcal{K}=\{0,\ldots,K\}$ the set of all early exits, by $N_k$ the size of the embedding at early exit $k$ (in bits), by $F_k$ the number of FLOPS that are needed to process data up to EE $k$. 
In the sequel, we formalize the KPIs including end-to-end loop delay and goal effectiveness.

\noindent \textbf{Communication and computation delay.} When sharing the inference computing workload among the device and the edge server, both local computing delay, communication delay to transfer intermediate results (when offloading) and remote computing delay (when offloading) must be taken into account. The end-to-end delay (or, loop delay) depends on the local computing resources, the choice of the EE, the choice on transmission, and the remote computing resources. We denote by $f_l$ (FLOPS/s) the computing power of the end device, and similarly by $f_r$ the CPU frequency of the edge server. 
Assuming that the device computes up to EE $k$ before either offloading or exiting, and assuming without loss of generality that $f_l$ is constant during the whole computation, we write the overall local computing delay as:
\begin{equation}
    D_{l,\text{comp}} = \frac{F_k}{f_l}.
\end{equation}
The communication delay depends on the choice of the EE, due to the size in bits of the data embedding, and the wireless channel state that determines the data rate (as a result of a channel state information and modulation and coding scheme (MCS) selection procedures). 
Denoting by $R$ the data rate, 
the transmission delay when offloading after EE $k$ reads as:
\begin{equation}
    D_{\text{tx}} = \frac{N_k}{R}.
\end{equation}
Finally, whenever the device offloads intermediate results to handover computation to the edge server, a remote computing delay should also be taken into account, depending on the computing resources of the hosting facility. When offloading after EE $k$ the latter can be written as follows:
\begin{equation}
    D_{r,\text{comp}}=\frac{F_K - F_k}{f_r}.
\end{equation}
Finally, the end-to-end loop delay reads as follows:
\begin{equation}
    D_{\text{tot}} = D_{l,\text{comp}} + D_{\text{tx}} + D_{r,\text{comp}}.
\end{equation}
$D_\text{tot}$ is a random variable influenced by channel conditions, compute resource availability at the device and edge server, as well as the EE and offloading decisions.

\noindent \textbf{Inference performance.} At inference time, accuracy cannot be directly assessed, necessitating proxy metrics for efficient communication and computation without compromising performance. This paper uses the margin (see Sec. II) as a proxy metric due to its online measurability during inference without ground truth and its availability after each EE. The only assumption is that accuracy correlates positively with the average margin, making it a reliable proxy for accuracy. However, we prioritize not just accuracy but timely inference within a set deadline, defining this KPI as goal-effectiveness \cite{merluzzi20246g}, i.e., 
\begin{equation}\label{go_eff}
    \mathcal{E}_{\text{go}} = \lim_{T\to\infty}\frac{1}{T}\sum\nolimits_{T=0}^{T-1} \mathbb{E}\{\mathbf{1}_{\{D_{\text{tot},T}\leq D_{\max}\}}\cdot\mathbf{1}_{\{A_T=1\}}\},
\end{equation}
where $\mathbf{1}_{\{P\}}$ is an indicator function that equals one if event $P$ occurs, and $A_T\in\{0,1\}$ is a binary variable that equals $1$ if data sample $A_T$ is correctly classified. 


\noindent \textbf{Online Early Exit decision via Reinforcement Learning.} We assume that time is organized in frames $T=1,2,\ldots$. At the beginning of each frame, a new data sample is available at the device and ready to be classified. Each frame is organized in $K$ slots $t=1,\ldots, K$, with $K$ being the total number of EEs. This is a design choice that helps us building a sequential decision making that lasts, at most, $K$ iterations.
Within a time frame, the collaborative inference process is modelled through a Markov chain and a corresponding Markov Decision Process (MDP). At time slot $t$, the state $s(t)$ is composed by two entries: i) the current EE $k(t)$, and ii) the current MCS that would be used in case of transmission during time slot $t$. The latter depends on the time varying wireless channel state. Then, denoting by $\mathcal{M}$ the set of MCSs, the state space has dimension $K\times \text{card}(\mathcal{M})$. At time $t=0$, i.e., at the beginning of the inference process for a new data sample, the state corresponds to EE $0$, coupled with the MCS offered by the physical layer (depending on channel state). 

At the start of each time slot, the end device selects one of the following three actions:
\begin{itemize}
    \item \textit{Action $a=0$}: Exit at the current EE and obtain the classification result.
    \item \textit{Action $a=1$}: Compute up to the next EE.
    \item \textit{Action $a=2$}: Offload intermediate results to hand over computation to the edge server.
\end{itemize}
Actions $a=0$ and $a=2$ directly lead to a terminal state, where the inference result is issued. Conversely, action $a=1$ progresses to the next EE, which could also result in a terminal state if full local computation is completed. An episode lasts at most an entire time frame, or it stops earlier if exit is selected. 

Our goal is to balance computational and communication burden and the goal's effectiveness.
Therefore, to define the reward at the end of an episode $T$, when exiting with either $a=0$ or $a=2$ (being at EE $k$), let us define the following KPIs related to communication and computation:
\begin{itemize}
    \item The \textit{relative computation workload saving} when exiting at EE $k$, which reads as follows: 
    \begin{equation}\label{comp_saving}
        \Gamma_{\text{comp}, k} = \frac{F_K-F_k}{F_K}.
    \end{equation}
    \item The \textit{relative payload size saving} when transmitting after EE $k$, which reads as follows: \begin{equation}\label{comm_saving}
        \Gamma_{\text{comm}, k} = \frac{\underset{j}{\max}(N_j) - N_k}{\underset{j}{\max}(N_j)}\mathbf{1}_{\{a_t^{(k)}=2\}} + \mathbf{1}_{\{a_t^{(k)}=0\}},
    \end{equation} 
    meaning that the maximum gain (i.e., $1$) is achieved when exiting locally, i.e., not offloading any computation.
    \item The output margin when exiting at EE $k$, denoted as $m_k$, coupled with the delay constraint, into the following \textit{proxy goal effectiveness metric}: 
    \begin{equation}\label{margin_th}
        \mathcal{E}_{\text{go},m} = \mathbf{1}_{\{D_{\text{tot},T}\leq D_{\max}\}}\cdot\mathbf{1}_{\{m_k>m_{\text{th}}\}},
    \end{equation}
    where $m_{\text{th}}$ is a predefined margin threshold.
\end{itemize}
Clearly, $0\leq \Gamma_{\text{comp}, k}\leq 1$, $0\leq \Gamma_{\text{comm}, k}\leq 1$, and $0\leq m_k\leq 1$, depending on the selected EE. 
We adopt a sparse reward structure, issued only at the end of an episode, which spans a maximum of $K$ rounds. The reward is triggered when the device opts to exit locally, offload, or complete full local computation. With $\gamma_{\text{comm}}$ ($\gamma_{\text{comp}}$) representing the weight for communication (computation) savings, the reward reads as:
\begin{align}\label{reward}
r_t = \begin{cases}
\gamma_{\text{comm}}\Gamma_{\text{comm, k}} +            \gamma_{\text{comp}}\Gamma_{\text{comp, k}}, \quad & \text{if}\; \mathcal{E}_{\text{go},m}=1,\\
-1, \; &\text{otherwise.}\;
\end{cases}
\end{align}
The reward in \eqref{reward} represents a weighted sum of communication and computation savings, provided both the delay and margin constraints are met. Otherwise, it incurs a significant penalty (-1) if either the delay exceeds the limit or the target margin (i.e., the proxy for inference accuracy) is not achieved. Finally, the online decision on the best action is  taken using a reinforcement approach based on Q-learning.


\section{Numerical results}

We consider a wireless scenario where a device is randomly placed within a circular area around an access point (AP), at distances between 10 m and 100 m. The AP communicates over a 20 MHz bandwidth at a carrier frequency of 3.5 GHz. The average path loss is given by $\text{PL}=\left(\frac{c}{4\pi f_c}\right)^2 d^{-\alpha}$ with a path loss exponent $\alpha=3.5$, combined with Rayleigh fading. The MCS is selected from $\mathcal{M}=\{0, 0.5, 1, 1.5, 2, 3, 4, 5\}$ bit/s/Hz, based on the highest value below the capacity $C=\log_2(1+\text{SNR})$, where SNR is derived from path loss, fading, and transmit power $P_{\text{tx}}=0.1$ W. The device can process the entire model in 50 ms, while the edge server completes it in 10 ms. Intermediate EE delays are proportional to the fraction of FLOPs required. Experiments use a delay threshold $D_{\max}=40$ ms and margin thresholds $m_{\text{th}} \in {0.09, 0.1, 0.2, 0.3}$. The computation weight is $\gamma_{\text{comp}}=1$, while the communication weight is varied as $\gamma_{\text{comm}} \in \{1, 1.5, 2, 2.5, 3, 3.5\}$.

\begin{figure}
\vspace{-.8cm}
    \centering    \includegraphics[width=.92\columnwidth]{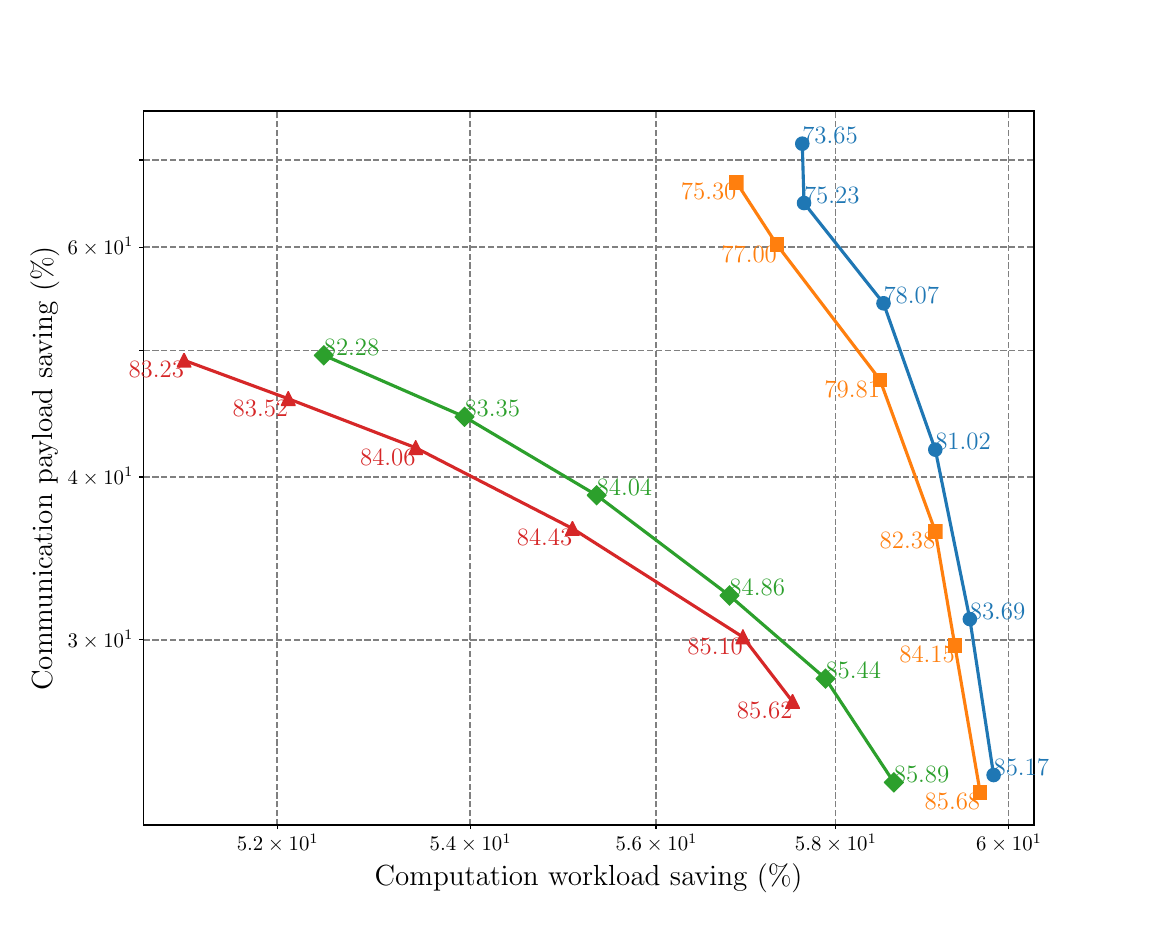}
    \caption{Communication-computation-effectiveness trade-off: the four curves are obtained via different margin thresholds}
    \label{fig:tradeoff}
\end{figure}

\begin{figure}[t]
\vspace{-.3cm}
    \centering
    \begin{subfigure}[a]{0.9\columnwidth}
        \includegraphics[width=\columnwidth]{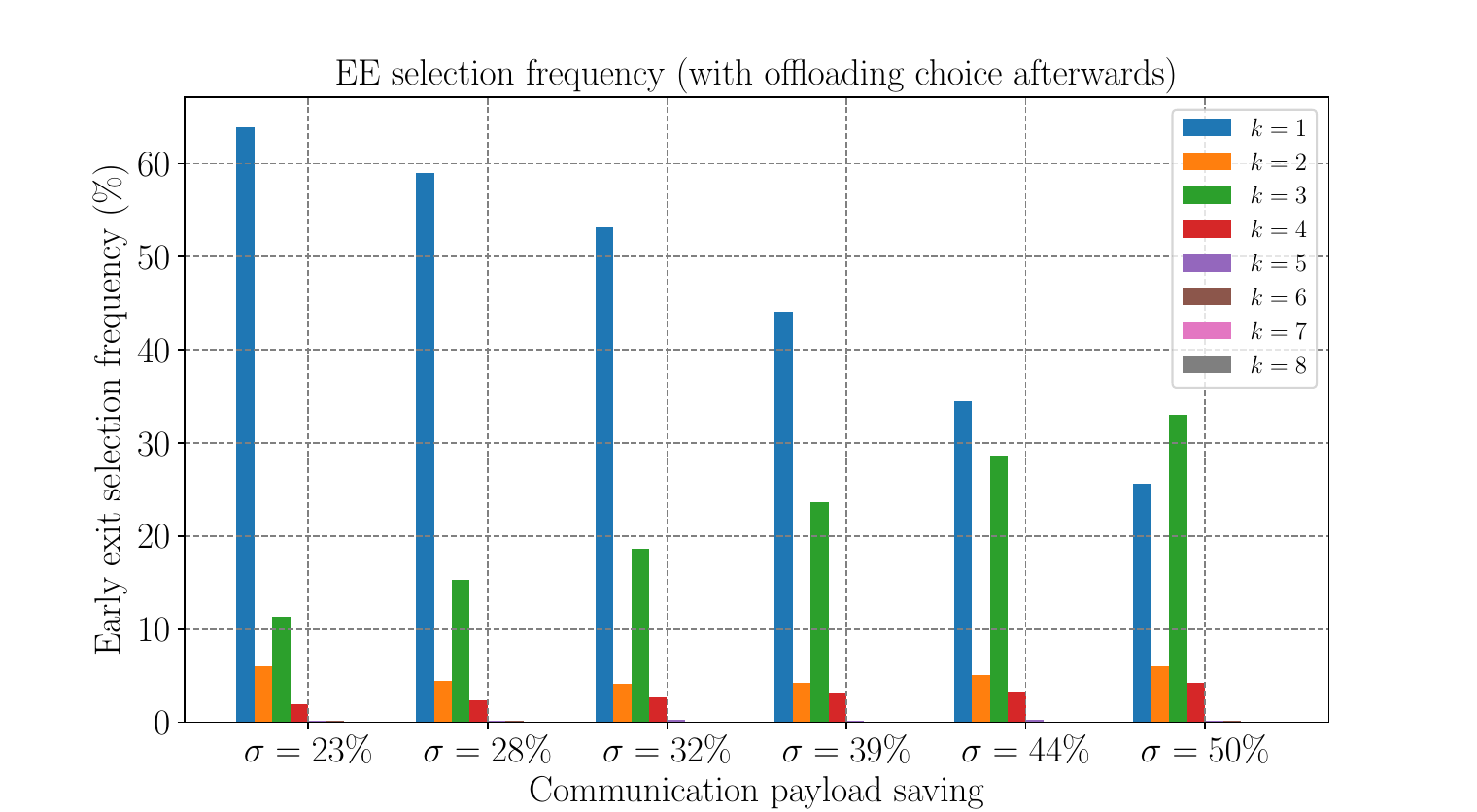}
        \caption{Offloading is performed after the selected EE.}
        \label{fig:ee_sel_off}
    \end{subfigure}
    
    \begin{subfigure}[a]{0.9\columnwidth}
        \includegraphics[width=\columnwidth]{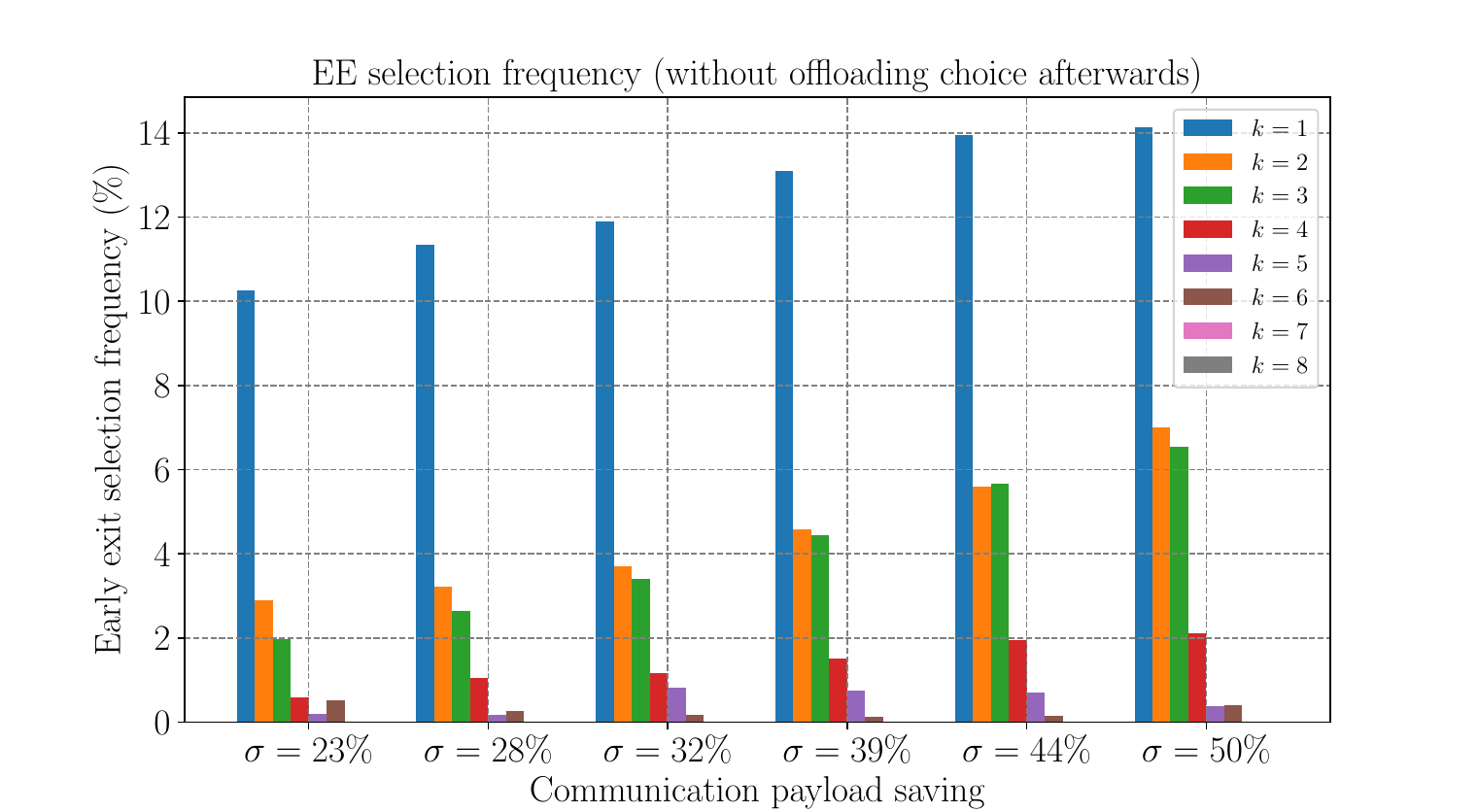}
        \caption{Offloading is not performed after the selected EE.}
        \label{fig:ee_sel_no_off}
    \end{subfigure}
    \caption{EE selection frequency vs. communication saving.}
    \label{fig:ee_sel}
    \vspace{-0.3 cm}
\end{figure}

To assess the performance of our approach, in 
Fig. \ref{fig:tradeoff}, we show the trade-off between computation workload saving (cf. \eqref{comp_saving}), communication payload saving (cf. \eqref{comm_saving}), and goal effectiveness (cf; \eqref{go_eff}).  The latter is shown as text for each point of the different curves. The colors represent different margin thresholds, with blue (RHS of the figure) indicating the lowest and red (LHS) the highest. Higher margin thresholds, which influence the reward in \eqref{reward}, drive the agent to learn solutions with greater goal effectiveness but at the expense of communication-computation savings. The blue curve demonstrates optimal communication-computation trade-offs, where a small loss in computation savings (x-axis) yields significant communication savings. However, this comes at the cost of reduced goal effectiveness. 
For each curve (representing a margin threshold), improving computation savings results in a trade-off with communication savings but leads to higher goal effectiveness. This is because more samples are offloaded earlier to the edge server, which allows the best performance. Specifically, this behaviour is visible in Figs. \ref{fig:ee_sel}a)-b), which show the frequency of each EE selection with and without offloading afterward, respectively, as a function of the achieved communication payload saving. These figures corresponds to a single point of Fig. \ref{fig:tradeoff}. From Figs. \ref{fig:ee_sel}a)-b), we observe that for the lowest communication savings ($23\%$), the device offloads over $60\%$ of data after the first EE. However, as communication saving requirements increase, deeper EE selection becomes preferable due to reduced data dimensionality from the third EE onward (green bar plot). Simultaneously, local exits are favored to conserve communication resources (cf. \ref{fig:ee_sel_no_off}), though this comes at the cost of increased computation or reduced computation savings. Overall, the method  leverages the flexibility offered by EEs and the dynamic nature of wireless channels to achieve an optimal balance between computation savings, communication savings, and goal effectiveness.

\section{Conclusions and future perspectives}

In this paper, we proposed a novel framework for goal-oriented semantic communications using recursive early exit models. Key contributions include an innovative early exit strategy for dynamic computation partitioning and a Reinforcement Learning-based optimization framework to jointly manage exit points, computation splitting, and offloading under varying wireless conditions. Numerical evaluations in edge inference scenarios confirm its effectiveness in balancing performance, latency, and resource efficiency. Several interesting research directions are open, including investigation of the multi-agent setting, and the implementation of mixed model-based and data-driven learning strategies for optimal allocation of computation and communication resources.

\bibliographystyle{./IEEEtran}
\bibliography{asilomar}

\begin{thebibliography}{10}
\providecommand{\url}[1]{#1}
\csname url@samestyle\endcsname
\providecommand{\newblock}{\relax}
\providecommand{\bibinfo}[2]{#2}
\providecommand{\BIBentrySTDinterwordspacing}{\spaceskip=0pt\relax}
\providecommand{\BIBentryALTinterwordstretchfactor}{4}
\providecommand{\BIBentryALTinterwordspacing}{\spaceskip=\fontdimen2\font plus
\BIBentryALTinterwordstretchfactor\fontdimen3\font minus \fontdimen4\font\relax}
\providecommand{\BIBforeignlanguage}[2]{{%
\expandafter\ifx\csname l@#1\endcsname\relax
\typeout{** WARNING: IEEEtran.bst: No hyphenation pattern has been}%
\typeout{** loaded for the language `#1'. Using the pattern for}%
\typeout{** the default language instead.}%
\else
\language=\csname l@#1\endcsname
\fi
#2}}
\providecommand{\BIBdecl}{\relax}
\BIBdecl

\bibitem{strinati2024goal}
E.~C. Strinati, P.~Di~Lorenzo, V.~Sciancalepore, A.~Aijaz, M.~Kountouris, D.~G{\"u}nd{\"u}z, P.~Popovski, M.~Sana, P.~A. Stavrou, B.~Soret \emph{et~al.}, ``Goal-oriented and semantic communication in 6g ai-native networks: The 6g-goals approach,'' \emph{arXiv preprint arXiv:2402.07573}, 2024.

\bibitem{di2023goal}
P.~Di~Lorenzo \emph{et~al.}, ``Goal-oriented communications for the {IoT}: System design and adaptive resource optimization,'' \emph{IEEE Internet of Things Magazine}, vol.~6, no.~4, pp. 26--32, 2023.

\bibitem{qu2020deep}
Z.~Qu, C.~Liu, and L.~Thiele, ``Deep partial updating: Towards communication efficient updating for on-device inference,'' \emph{arXiv preprint arXiv:2007.03071}, 2020.

\bibitem{8763885}
J.~Chen and X.~Ran, ``Deep learning with edge computing: A review,'' \emph{Proceedings of the IEEE}, vol. 107, no.~8, pp. 1655--1674, 2019.

\bibitem{9116699}
A.~F. Rocha~Neto, F.~C. Delicato, T.~V. Batista, and P.~F. Pires, \emph{Dist. Machine Learning for IoT Appl. in the Fog}, 2020, pp. 309--345.

\bibitem{MerluzziHexa23}
M.~Merluzzi \emph{et~al.}, ``The hexa-x project vision on artificial intelligence and machine learning-driven communication and computation co-design for 6g,'' \emph{IEEE Access}, vol.~11, pp. 65\,620--65\,648, 2023.

\bibitem{Ana21}
P.~M. de~Sant~Ana, N.~Marchenko, P.~Popovski, and B.~Soret, ``Age of loop for wireless networked control systems optimization,'' in \emph{2021 IEEE 32nd Annual International Symposium on Personal, Indoor and Mobile Radio Communications (PIMRC)}, 2021, pp. 1--7.

\bibitem{9700550}
E.~Samikwa, A.~Di~Maio, and T.~Braun, ``Adaptive early exit of computation for energy-efficient and low-latency machine learning over iot networks,'' in \emph{IEEE CCNC}, 2022, pp. 200--206.

\bibitem{Binucci24}
F.~Binucci, M.~Merluzzi, P.~Banelli, E.~C. Strinati, and P.~Di~Lorenzo, ``Enabling edge artificial intelligence via goal-oriented deep neural network splitting,'' in \emph{Proc. of ISWCS}, 2024, pp. 1--6.

\bibitem{teerapittayanon2016branchynet}
S.~Teerapittayanon, B.~McDanel, and H.-T. Kung, ``Branchynet: Fast inference via early exiting from deep neural networks,'' in \emph{Proc. of ICPR}.\hskip 1em plus 0.5em minus 0.4em\relax IEEE, 2016, pp. 2464--2469.

\bibitem{Bullo23}
M.~Bullo, S.~Jardak, P.~Carnelli, and D.~Gündüz, ``Sustainable edge intelligence through energy-aware early exiting,'' in \emph{Proc. of IEEE MLSP}, 2023, pp. 1--6.

\bibitem{Jankowski24}
M.~Jankowski, D.~Gündüz, and K.~Mikolajczyk, ``Adaptive early exiting for collaborative inference over noisy wireless channels,'' in \emph{Proc. of IEEE ICMLCN}, 2024, pp. 126--131.

\bibitem{Colocrese24}
H.~S. M.~Colocrese, E.~Koyuncu, ``Early-exit meets model-distributed inference at edge networks,'' \emph{arXiv preprint: https://www.arxiv.org/abs/2408.05247}, 2024.

\bibitem{probreint}
J.~Pomponi, S.~Scardapane, and A.~Uncini, ``A probabilistic re-intepretation of confidence scores in multi-exit models,'' \emph{Entropy}, 2022.

\bibitem{wolczyk2021zero}
M.~Wo{\l}czyk, B.~W{\'o}jcik, K.~Ba{\l}azy, I.~T. Podolak, J.~Tabor, M.~{\'S}mieja, and T.~Trzcinski, ``Zero time waste: Recycling predictions in early exit neural networks,'' \emph{Proc. of NeurIPS}, vol.~34, pp. 2516--2528, 2021.

\bibitem{he2016deep}
K.~He, X.~Zhang, S.~Ren, and J.~Sun, ``Deep residual learning for image recognition,'' in \emph{Proceedings of the IEEE conference on computer vision and pattern recognition}, 2016, pp. 770--778.

\bibitem{zhou2020bert}
W.~Zhou, C.~Xu, T.~Ge, J.~McAuley, K.~Xu, and F.~Wei, ``Bert loses patience: Fast and robust inference with early exit,'' \emph{Advances in Neural Information Processing Systems}, vol.~33, pp. 18\,330--18\,341, 2020.

\bibitem{merluzzi20246g}
M.~Merluzzi \emph{et~al.}, ``6g goal-oriented communications: How to coexist with legacy systems?'' in \emph{Telecom}, vol.~5, no.~1, 2024, pp. 65--97.

\end{thebibliography}
\end{document}